\documentclass[letterpaper, 10 pt, conference]{ieeeconf}  % Comment this line out if you need a4paper

\IEEEoverridecommandlockouts                              % This command is only needed if 
\usepackage{amsmath,amssymb,amsfonts}
\usepackage{graphicx}
\usepackage{comment}
\usepackage{amsmath}
\usepackage{amssymb}
\usepackage{caption}
\usepackage{subcaption}
\usepackage{soul}
\usepackage{mathtools}
\usepackage{import}
\usepackage[nocompress]{cite}

\graphicspath{ {./images/} }

\usepackage[x11names, xcdraw, table]{xcolor}
\usepackage{booktabs}
\usepackage{etoolbox}
\usepackage{pgf}

\definecolor{high}{HTML}{f3acac}
\definecolor{low}{HTML}{e9f2fa}
\definecolor{top}{HTML}{ea667a}
\newcommand*{\opacity}{90}% here you can change the opacity of the background color!

\usepackage{xcolor}
\definecolor{ourblue}{rgb}{0.368,0.507,0.71}
\definecolor{ourgreen}{rgb}{0.56,0.692,0.195}
\definecolor{ourviolet}{rgb}{0.528,0.471,0.701}

\def\eqref#1{(\ref{#1})}

\newcommand{\eg}{\emph{e.g.},{ }}

\newcommand{\psib}{\Psi_{\mathrm{b}}}

\newcommand{\pe}{p_{\mathrm{e}}}

\newcommand{\fb}{\{\mathrm{b}\}}

\newcommand{\e}{\mathrm{e}}
\newcommand{\B}{\mathrm{B}}
\newcommand{\R}{\mathbb{R}}
\newcommand{\PRB}{\mathrm{prb}}
\newcommand{\PRBN}{\text{PRB-Net}}
\newcommand{\nseg}{n_{\text{el}}}

\newcommand*{\minval}{0.0}% define the minimum value on your data set
\newcommand*{\maxval}{100.0}% define the maximum value in your data set!
\newcommand{\gr}[1]{
    \ifdimcomp{#1pt}{>}{\maxval pt}{#1}{
        \ifdimcomp{#1pt}{<}{\minval pt}{#1}{
            \pgfmathparse{int(round(100*(#1/(\maxval-\minval))-(\minval*(100/(\maxval-\minval)))))}
            \xdef\tempa{\pgfmathresult}
            \cellcolor{high!\tempa!low!\opacity} #1
    }}
}
\newcommand{\grc}[6]{
    \ifdimcomp{#1pt}{>}{#3 pt}{#1}{
        \ifdimcomp{#1pt}{<}{#2 pt}{#1}{
            \pgfmathparse{int(round(100*(#1/(#3-#2))-(\minval*(100/(#3-#2)))))}
            \xdef\tempa{\pgfmathresult}
            \cellcolor{#5!\tempa!#4!#6} #1
    }}
}

\newcommand*{\minvals}{0.0}% define the minimum value on your data set
\newcommand*{\maxvals}{100.0}% define the maximum value in your data set!
\newcommand{\grs}[1]{
    \ifdimcomp{#1pt}{>}{\maxvals pt}{#1}{
        \ifdimcomp{#1pt}{<}{\minvals pt}{#1}{
            \pgfmathparse{int(round(100*(#1/(\maxvals-\minvals))-(\minvals*(100/(\maxvals-\minvals)))))}
            \xdef\tempa{\pgfmathresult}
            \cellcolor{high!\tempa!low!\opacity} #1
    }}
}
\title{\LARGE \bf 
Pseudo-rigid body networks: learning interpretable deformable\\ object dynamics from partial observations
}

\author{Shamil Mamedov$^{1,\ast}$, A.\ René Geist$^{2,\ast}$, Jan Swevers$^{1}$, Sebastian Trimpe$^{2}$% <-this % stops a space
\thanks{$^\ast$ \textbf{Equal contribution.}}
\thanks{This work was largely done during an exchange stay of S. Mamedov at RWTH Aachen University, which was supported by the FWO-Vlaanderen through SBO project ELYSA for cobot applications (S001821N).}% <-this % stops a space
\thanks{$^1$MECO Research Team, KU Leuven and Flanders Make@KU Leuven. }
\thanks{$^2$Institute for Data Science in Mechanical Engineering, RWTH Aachen
University.}%
}
\makeatletter
\let\NAT@parse\undefined
\makeatother
\usepackage[hidelinks]{hyperref}
\usepackage[switch]{lineno}
\hypersetup{
	colorlinks=true,       % false: boxed links; true: colored links
	linkcolor=ourblue,        % color of internal links
	citecolor=ourviolet,        % color of links to bibliography
	filecolor=ourblue,     % color of file links
	urlcolor=ourviolet
}

\begin{document}

\maketitle
\thispagestyle{empty}
\pagestyle{empty}

%%%%%%%%%%%%%%%%%%%%%%%%%%%%%%%%%%%%%%%%%%%%%%%%%%%%%%%%%%%%%%%%%%%%%%%%%%%%%%%%
\begin{abstract}
    Accurately predicting deformable linear object (DLO) dynamics is challenging, especially when the task requires a model that is both human-interpretable and computationally efficient. In this work, we draw inspiration from the pseudo-rigid body method (PRB) and model a DLO as a serial chain of rigid bodies whose internal state is unrolled through time by a dynamics network. This dynamics network is trained jointly with a physics-informed encoder that maps observed motion variables to the DLO's hidden state. To encourage the state to acquire a physically meaningful representation, we leverage the forward kinematics of the PRB model as a decoder. We demonstrate in robot experiments that the proposed DLO dynamics model provides physically interpretable predictions from partial observations while being on par with black-box models regarding prediction accuracy. The project code is available at:  \href{http://tinyurl.com/prb-networks}{tinyurl.com/prb-networks}
    %\url{https://tinyurl.com/fei-networks}
\end{abstract}

%%%%%%%%%%%%%%%%%%%%%%%%%%%%%%%%%%%%%%%%%%%%%%%%%%%%%%%%%%%%%%%%%%%%%%%%%%%%%%%%
% \vspace{-2mm}

\section{Introduction}

% \vspace{-1mm}

Deformable linear objects (DLOs), such as cables, ropes, and threads, are prevalent in various promising applications within the field of robotics \cite{sanchez2018robotic, jimenez2012survey}. Dynamics models empower robots to interact with DLOs, showcasing remarkable dexterity at tasks such as intricate knotting \cite{wang2015online, yoshida2015simulation}, precise manipulation of ropes \cite{yan2020selfsupervised}, and surgical suturing \cite{pai2002strands, saha2007manipulation}. 
In addition, the parts of many robots can be modeled as DLOs. This naturally applies to numerous soft robots but is also relevant for controlling industrial robots \cite{verl2019}. 
%For example, the significant end-effector forces arising in robot machining cause elastic deformations which, if not modeled accurately, reduce manufacturing precision . In addition, designing lightweight yet flexible manipulators and legged robots potentially lowers the robot's manufacturing costs, increases its maximum attainable joint accelerations, and benefits its energy efficiency \cite{drucker2023application}.

Learning robot dynamics is challenging because the amount of real data is often limited, and the data's coverage of the robot’s state space depends on the control strategy used for data collection. Therefore, purely data-driven models, such as neural networks (NNs), often fall short of expectations when learning robot dynamics \cite{sanchez2018robotic, jimenez2012survey}. Furthermore, these models lack interpretability. 
A promising approach to enhancing both sample efficiency and interpretability is to incorporate first-principles knowledge  into the model's architecture \cite{lutter2019deep, cranmer2019lagrangian}. 
Physics-informed models are generally evaluated on rigid robots whose states are directly observed \cite{rath2022using, nguyen2010using, lutter2023combining}.
Unlike rigid robots, a DLO is a continuous accumulation of mass, making it unclear what the state representation should be and how to estimate it in the first place. 
Fortunately, we can draw inspiration from the literature on the pseudo-rigid body (PRB) method and the articulated body algorithm to pursue the simple idea:
\begin{quote}
    \emph{Model a DLO as a chain of rigid bodies whose dynamics are learned from data.}
\end{quote}
This approach combines the strengths of the PRB method and machine learning. 
First, it provides a physically interpretable state.
%that pertains to the degrees of freedom (DOF) between the bodies.
Second, it infers the interactions between the bodies from data instead of solely relying on the expressiveness of analytical physics models.
%On the other hand, analytically modeling interactions between those bodies becomes difficult, particularly for coarse PRB discretizations with few rigid bodies.
%By learning these interactions, we obtain a computationally fast and accurate model suitable, e.g. for model-based control of DLOs.

\begin{figure}[t]
    \captionsetup{font=small}
    \centering
    \includegraphics[width=0.88\columnwidth, trim={8.6cm 0 0 0}, clip]{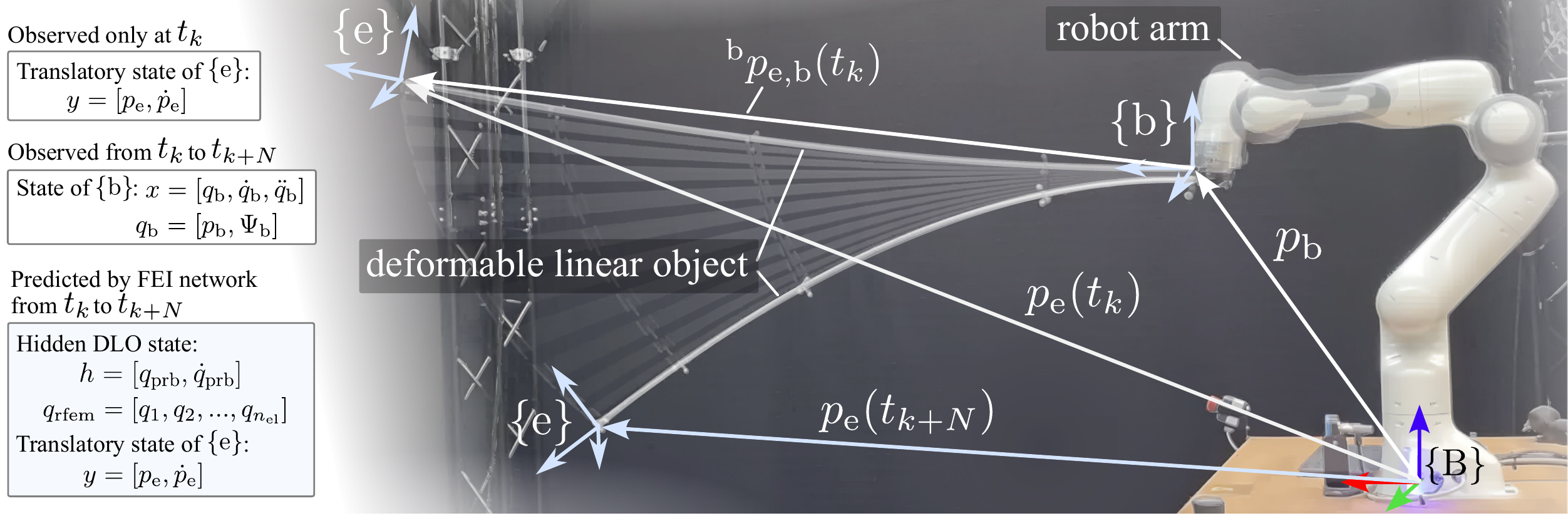}
    \caption{A DLO -- an aluminium rod -- is actuated by a robot arm. Given an initial observation of $\{\mathrm{e}\}$'s translational state $y_k$ and $\{\mathrm{b}\}$'s position, velocity and acceleration vector  $x_{k}$, a \PRBN~predicts $y_{k+1}$ while also estimating the DLO's hidden state $h_k$.
    }
    \label{fig:setup_and_method}
    \vspace{-7mm}
\end{figure}

\vspace{-0.5mm}

\subsection{Problem Statement and Contributions}
%In short, this work solves the following problem: \textit{given measurements of positions, velocities and accelerations of one end of a DLO, predict the position and velocity of its other end over time}. 
%Ultimately, we aim to use such a model for solving DLO manipulation tasks, thus we additionally prioritize models that are computationally fast and can be used within model predictive controllers. 

Figure~\ref{fig:setup_and_method} gives an overview of the problem setup. The inertial coordinate frame reads $\{\mathrm{B}\}$ while $\{\text{b}\}$ and $\{\text{e}\}$ denote frames fixed to the DLO's start and end, respectively. The DLO's hidden state is $ h\in \R^{n_h}$. A dataset $\mathcal{D}$ contains trajectories consisting of the position and velocity of $\{\text{e}\}$ relative to $\{\mathrm{B}\}$, denoted by $y=[p_\text{e}, \dot p_\text{e}] \in \R^{n_y}$ and position, velocity, and acceleration of $\{\text{b}\}$ relative to $\{\mathrm{B}\}$, denoted as $x=[q_\text{b}, \dot q_\text{b}, \ddot q_\text{b}] \in \R^{n_x}$. %Notably, all variables are expressed in the $\{B\}$ frame. 
In short, we tackle the problem: \textit{given measurements of position, velocity, and acceleration of a DLO's end, predict position and velocity of its other end over the next $N$ time steps} or, using our notation:
\begin{quote}
\hspace{-5mm} \emph{given $y_k$ and $x_{k},...\,, x_{k+N-1}$ predict $y_{k+1},...\,, y_{k+N}$, }
\end{quote}
where $k$ is the discrete time index (or step).
\noindent To this end, we propose the PRB network (PRB-Net) which consists of four essential components: 1) a physics-informed encoder that maps  $y_k$ and $x_k$ to $h_k$, 2) a dynamics network that predicts $h_{k+1}$ from $h_k$ and $x_k$, 3) a decoder that maps $h_k$ and $x_k$ to $y_k$, and 4) a loss function that regularizes $h$ during training.
The contributions of this work are:
\begin{enumerate}%[label={\textit{\roman*})}] 
\itemsep0em 
    \item We introduce \PRBN, a model designed to predict the dynamics of DLOs from partial observations.
    \item We demonstrate that using the forward kinematics of a DLO's PRB discretization as a decoder effectively enforces a physically consistent hidden state.
    \item Through experiments on two different DLOs, we show that \PRBN's predictions are on par with black-box models in terms of accuracy and computation times, while being physically interpretable. 
\end{enumerate}

%============================================================================================

\section{Related Work} \label{sec:related_work}
The main contribution of our work revolves around learning hidden state dynamics of a DLO from partial observations using PRBs. Therefore, special emphasis is placed on the type of observations that are used by the different DLO dynamics models.

\paragraph{Analytical physics-based models}
Numerous works explore physics-based dynamics modeling of DLOs \cite{spillmann2007corde, drucker2023application, kim2011fast}. An in-depth discussion of the mechanics underlying DLO is given in \cite{bergou2008discrete}.
The dynamics of DLOs can be conceptualized as an aggregation of infinitesimally small mass particles interacting through forces. 
To render the complicated particle interactions tractable for computational analysis, spatial discretization methods aggregate the continuous distribution of particles into finite elements. 
The PRB method assumes that the particles inside an element form a rigid body such that their distances to each other remain fixed \cite{wittbrodt2007RFEM}. 
The PRB~approximation results in a simplified representation of the DLO's dynamics, but its accuracy depends on the chosen force function that determines element interactions and its parameters. 
In particular, the oftenly used linear spring-damper force models are inaccurate for coarse spatial discretizations or modeling DLOs with heterogeneous material properties.
% As PRB lies at the heart of our work, we discuss it in more detail in Section \ref{sec:fem}.

Cosserat rods represents an alternative method for modeling DLOs, particularly favored by the soft robotics community. Although these models are often quite accurate and geometrically exact, their large prediction times due to numerical integration renders them impractical for model-based control\cite{boyer2024CosseratRodInteg}. To mitigate the computational demands of Cosserat rod models, finite difference \cite{lang2011multi, bergou2008discrete} and piecewise constant curvature \cite{webster2010design} approximations have been proposed.
Stella et al.\ employed the latter to model the dynamics of a 2D underwater tentacle \cite{stella2022underwatertentacle} and the kinematics of a 3D soft-continuum robot  \cite{stella2022piecewise}.

This paper adopts the state-representation and kinematics from the analytical methods, specifically from PRB method. Instead of depending on analytical dynamics, which require numerical integration, struggle to capture certain real-world phenomena, and are computationally unsuitable for control, we learn its discrete-time approximation from data. 

\begin{table}[t]
\captionsetup{font=small}
\centering
\footnotesize
%\scriptsize
%\renewcommand{\arraystretch}{1.2}
{\setlength{\tabcolsep}{3.3pt}
\def\arraystretch{1.2}
\begin{tabular}{lllll}
\toprule
\textbf{} & \textbf{System} & \textbf{Dynamics model} & \textbf{State type} & \textbf{Observation} \\ \midrule
\rowcolor[HTML]{ECF4FF}
\textbf{\cite{stella2022underwatertentacle}} & \begin{tabular}[c]{@{}l@{}}Submerged \\ tentacle (2D)\end{tabular} & \begin{tabular}[c]{@{}l@{}}constant \\curvature\end{tabular} & Splines & Full state \\
\textbf{\cite{tanaka2022continuum-body-pose}} & \begin{tabular}[c]{@{}l@{}}Submerged \\ tentacle (2D)\end{tabular} & \begin{tabular}[c]{@{}l@{}}echo state \\ network\end{tabular} & Points & Full state \\
\rowcolor[HTML]{ECF4FF}
\textbf{\cite{tariverdi2021recurrent}} & \begin{tabular}[c]{@{}l@{}}DLO in \\ magnetic field\end{tabular} & LSTMs & Points & Full state \\
\textbf{\cite{yan2020selfsupervised}} & \begin{tabular}[c]{@{}l@{}}Rope moved \\ by robot arm\end{tabular} & bi-LSTM & \begin{tabular}[c]{@{}l@{}}Mass \\ particles\end{tabular} & Full state \\
\rowcolor[HTML]{ECF4FF}
\textbf{\cite{yang2021learning}} & \begin{tabular}[c]{@{}l@{}}DLO attached \\ to robot arm\end{tabular} & \begin{tabular}[c]{@{}l@{}}bi-LSTM + \\ Graph neural network\end{tabular} & \begin{tabular}[c]{@{}l@{}}Mass \\ particles\end{tabular} & Full state \\
\textbf{\cite{preiss2022trackingfast}} & \begin{tabular}[c]{@{}l@{}}DLO attached \\ to robot arm\end{tabular} & LSTM & \begin{tabular}[c]{@{}l@{}}Black \\ box\end{tabular} & \begin{tabular}[c]{@{}l@{}}DLO \\ ends\end{tabular} \\
\rowcolor[HTML]{ECF4FF}
% \textbf{Ours} & \begin{tabular}[c]{@{}l@{}}DLO attached \\ to robot arm\end{tabular} & flexible(\emph{\eg RNN, ResNet}) & \begin{tabular}[c]{@{}l@{}}Rigid \\ bodies\end{tabular} & \begin{tabular}[c]{@{}l@{}}DLO\\ ends\end{tabular} \\ \bottomrule
\textbf{Ours} & \begin{tabular}[c]{@{}l@{}}DLO attached \\ to robot arm\end{tabular} & \begin{tabular}[c]{@{}l@{}}flexible \\ (\emph{\eg RNN, ResNet})\end{tabular}& \begin{tabular}[c]{@{}l@{}}Rigid \\ bodies\end{tabular} & \begin{tabular}[c]{@{}l@{}}DLO\\ ends\end{tabular} \\ \bottomrule
\end{tabular}}
\normalsize
%\vspace{1mm}
\caption{Overview of related literature on data-driven modeling of DLO dynamics. While numerous works focus on the dynamics identification and observe the DLO's full state through cameras, \cite{preiss2022trackingfast} and our work learn the dynamics by solely observing frames attached to the DLO's start and end. Moreover, while previous literature focuses on a particular model architecture to learn dynamics, we focus on leveraging kinematics such that \emph{any} dynamics model learns a physically meaningful representation.}
\label{tab:overview_on_literature}
\vspace{-5mm}
\end{table}

\paragraph{Data-driven models}
Table \ref{tab:overview_on_literature} presents a summary of related data-driven DLO dynamics models. Our research is inspired by \cite{preiss2022trackingfast}, which employed a long short-term memory (LSTM) network \cite{hochreiter1997lstm} to capture the input-output dynamics of a robot-arm-held foam cylinder. In this study, the authors utilize the LSTM to predict the DLO's end positions based on the pitch and yaw angles of its other end.
The LSTM's hidden state lacks a physically meaningful representation hindering the formulation of constraints on the DLO's shape, particularly during manipulation in an environment with obstacles. Additionally, prediction errors of a black-box model can quickly accumulate causing the trajectory prediction to become unstable. PRB-Net alleviates these limitations by resorting to kinematics.

In related works, \cite{tanaka2022continuum-body-pose, tariverdi2021recurrent} employed camera-based techniques to estimate the DLO's center-line as a sequence of points. 
In \cite{tariverdi2021recurrent}, the dynamics of each point are predicted by an LSTM that shares with the other LSTMs its output history. 
Similarly, \cite{tanaka2022continuum-body-pose} introduced echo state networks, where the output of an RNN serves as the input for a second RNN, in addition to control inputs and bending sensor measurements. 
In \cite{yan2020selfsupervised}, the authors developed a quasi-static model of a DLO using a bi-directional LSTM (bi-LSTM) whose state is estimated from camera images using a convolutional neural network. 
Unlike previous works that used LSTMs to propagate states in time, the bi-LSTM propagates information along both directions of the DLO's spatial dimension. 
Inspired by \cite{yan2020selfsupervised}, \cite{yang2021learning} combined bi-LSTMs with graph neural networks \cite{scarselli2008graph}. In particular, \cite{yang2021learning} discretizes the DLO into a sequence of cylindrical elements similar to Cosserat theory \cite{pai2002strands} while the spatial and temporal element interactions are modeled through the NNs. However, these seminal works require cameras to obtain an estimation of several points on the DLO and operate at low velocities where quasi-static models are sufficient for predicting states. In contrast, the proposed \PRBN~allows for extending these methods to operate on partial observations and in highly dynamic scenarios. In particular, the network in \PRBN~could be replaced by bi-LSTMs and graph neural networks.

\begin{comment}

Numerous other works explore learning dynamics purely in simulation. As these work do not encounter the many challenges of real-world robotics, they can focus solely on the algorithmic-side of combining physics with machine learning.
In this domain, one often encounters the ``Encoder-Processor-Decoder'' network structure that we also adopt in the FEI network. For example, \cite{sanchez2020learning} uses this structure to learn the dynamics of particle systems via GNNs. In \cite{sanchez2020learning}, all particle positons are observed and the model predicts acceleration which is then numerically integrated. In comparison, the FEI network uses the encoder as a state observer while its processor (aka ``dynamics network'') simulates the DLO's hidden discrete-time dynamics. 
In turn, one could further extend the FEI network by replacing its dynamics network with a GNN as proposed in \cite{sanchez2020learning}.
\end{comment}

\vspace{2mm}

%==============================================================================================
\section{Background: PRB method for modeling DLOs}
\label{sec:fem}
The PRB method approximates a DLO as a chain of $\nseg+1$ rigid bodies connected via elastic joints. \PRBN~explicitly uses forward kinematics of a DLO's PRB discretization and uses the same input-output structure as found in the dynamics of rigid-body chains. The following subsections briefly describe the kinematics and dynamics of a DLO using the PRB approximation.

\vspace{2mm}

\subsection{Kinematics} \label{subsec: dlo_kin}
\begin{figure}[t]
    \captionsetup{font=small}
    \centering
    \includegraphics[width=0.425\textwidth]{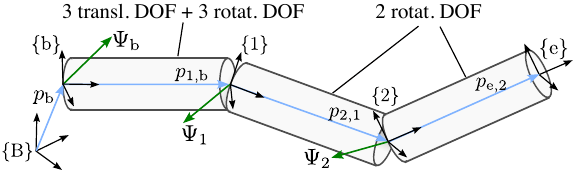}
    \caption{Generalized coordinate representations of a body chain with $\nseg=2$. The chain's joint states are described through rotation parameters $\Psi_i$ that provide two rotational DOF to each element. %\st{Alternatively, one could resort to using the distance vectors $p_{i+1,i}$.}
    }
    \label{fig:state-representation}
    \vspace{-6mm}
\end{figure}

%For the sake of a compact presentation, we use homogeneous transformation matrices and the spatial vector notation as detailed in \cite{featherstone2014rigidbody, kim2012lie, lynch2017modernrobotics}.
%
As illustrated in Fig. \ref{fig:state-representation}, the pose of the $i$-th body in the chain is described via a body-fixed coordinate frame $\{i\}$. The pose of $\{\e\}$ with respect to $\{\B\}$ is described by the homogeneous transformation matrix 
\begin{equation} \label{eq:transfo-chain}
    {}^{\B}T_{\e} = 
    \begin{bmatrix} {}^{\B}R_{\e} & {}^{\B}p_{\e,\B} \\ 0 & 1\end{bmatrix}\in \text{SE}(3) \nonumber
\end{equation}
with the position vector ${}^{\B}p_{\e,\B} \in \mathbb{R}^3$, rotation matrix ${}^{\B}R_{\e}$, and $\text{SE}(3)$ denoting the special Euclidean group \cite{lynch2017modernrobotics}. Note that we omit the superscripts and subscripts if a vector or matrix is either expressed in the inertia frame $\{B\}$ or it points from $\{B\}$'s origin such that $p_{\e}:={}^{\B}p_{\e,\B}$ and $R_{\text{e}}:={}^{\B}R_{\e}$. %whose elements perform first rotation and then translation of right-hand frames in Cartesian space. 
In a chain of $\nseg+1$ bodies ${}^{\text{b}} p_{\text{e},\text{b}}$ is obtained as 
\begin{equation}
    \begin{bmatrix} {}^{\mathrm{b}}p_{\mathrm{e}, \mathrm{b}} \\ 1 \end{bmatrix} = {}^{\mathrm{b}}T_{1}(q_1) \cdot {}^{1}T_{2}(q_2)  \cdot ...  \cdot {}^{\nseg-1}T_{\nseg}(q_{\nseg})  \begin{bmatrix}{}^{\nseg}p_{\e,\nseg} \\ 1 \end{bmatrix}, \nonumber %\label{eq:transfo_chain}
\end{equation}
where $q_\PRB = [q_1, ... , q_{\nseg}]$ denotes the generalized coordinates of the DLO's rigid body chain approximation and ${}^{i-1}T_{i}(q_i)$ transforms a (homogeneous) vector from $\{i\}$ to $\{i-1\}$ \cite[p.\,93]{lynch2017modernrobotics}. In addition, $q_{\mathrm{b}}=[p_{\mathrm{b}}, \psib]$ defines the six degrees of freedoms (DOF) of the floating-base $\fb$. 

For simplicity's sake, we employ two DOF revolute joints as illustrated in Fig. \ref{fig:state-representation} that are parameterized by Euler angles. Note that any other joint parametrization could be used such as free joints that provide six DOFs. In particular, three dimensional rotational joints should be used if the DLO is subject to torsion. In this case, recent works on learning with 3D rotations advocate against using Euler angles for large changes in rotation \cite{huynh2009metrics, bregier2021deep}.
% Various parametrizations $\Psi$ exist that describe rotational DOFs. Typically used rotation descriptions are Euler angles, quaternions, exponential coordinates \cite{lynch2017modernrobotics}, and position-twist angle pairs \cite{lefevre20174}. 

Using the forward kinematics of the PRB approximation, the DLO's endpoint position is obtained as follows
\begin{equation}
    p_{\e} = \mathrm{fk}(q_{\mathrm{b}}, q_\PRB) \coloneqq  R_{\text{b}}(\Psi_{\mathrm{b}}) \, {}^{\mathrm{b}} p_{\mathrm{e},\mathrm{b}}(q_\PRB) + p_{\mathrm{b}}. \label{eq:difpos}
\end{equation}
The velocity the DLO's end is obtained from differential kinematics which describes the relation between $\dot q_{\mathrm{b}}$, $ \dot q_\PRB$, and  $\dot p_{\mathrm{e}}$ via the joint Jacobian matrix $J_{\text{lin}}(q_\mathrm{b}, q_\PRB)$ as
\begin{equation}
    % {}^{\text{B}} \dot p_{\text{e}, \text{b}} = J_{\text{lin}}(q) \dot q = \sum_{i=1}^{\nseg} \frac{\partial {}^{\text{B}} p_{\text{e}, \text{b}}}{\partial q_i} \dot q_i. 
    \dot p_{\text{e}} = J_{\text{lin}}(q_\text{b}, q_\PRB) 
    \begin{bmatrix}
        \dot q_{\text{b}} \\ \dot q_\PRB
    \end{bmatrix} = 
    \begin{bmatrix}
        \frac{\partial \text{fk}}{\partial q_{\text{b}}}  & \frac{\partial \text{fk}}{\partial q_\PRB} 
    \end{bmatrix}
    \begin{bmatrix}
        \dot q_{\text{b}} \\ \dot q_\PRB
    \end{bmatrix}. \label{eq:difkin}
\end{equation}
Equation \eqref{eq:difpos} and \eqref{eq:difkin} together form the system's \emph{forward kinematics} (FK) that maps elements in $h_k$ and $x_k$ to $y_k$.

\begin{figure*}[t]
    \captionsetup{font=small}
    \centering
    \includegraphics[width=0.95\textwidth]{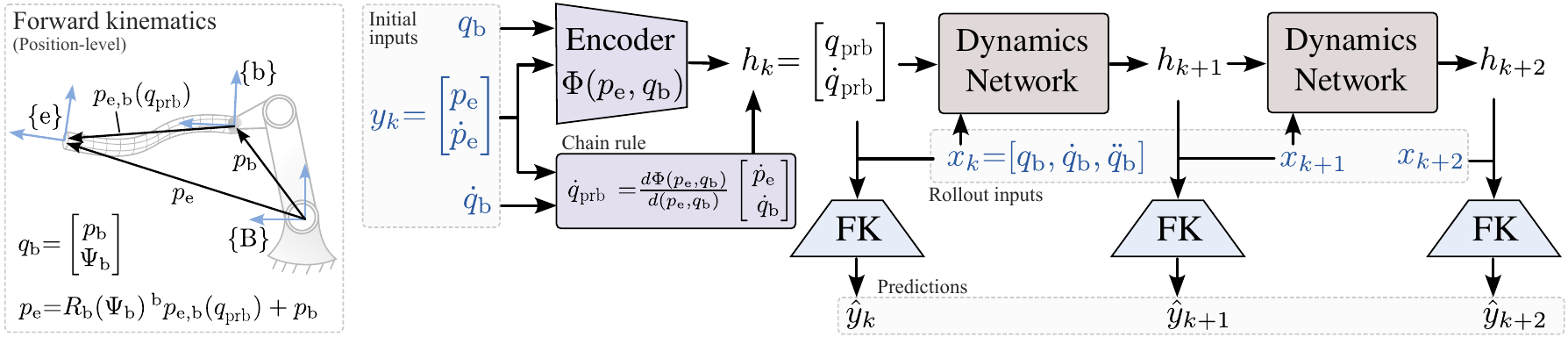}
    \caption{Architecture of PRB-Net with rollout length $N=2$. By resorting to the forward kinematics (FK) of a chain of PRBs and taking inspiration from hybrid dynamics to determine the inputs, the model learns a physically-plausible hidden state from partial observations. %that is unrolled through time.
    }
    \label{fig:FEIN}
    \vspace{-3mm}
\end{figure*}

\subsection{Dynamics}
The dynamics of a pseudo-rigid body chain can be categorized into forward, inverse, or hybrid dynamics. Forward dynamics uses joint torques as inputs to determine joint accelerations, while inverse dynamics calculates joint torques from known joint accelerations.
The dynamics of a DLO being manipulated by a robot arm is governed by \emph{hybrid dynamics} being the combination of forward and inverse dynamics \cite{kim2012lie}. For the first body attached to $\{\mathrm{b}\}$, the acceleration $\ddot q_b$ is given and $\tau_{b}$ is unobserved. For the other bodies, $\ddot q_{\PRB}$ is a function of $\tau_{\PRB}(q_{\PRB}, \dot q_{\PRB})$.
% Then, we wish to predict the accelerations $\ddot q_{\PRB}$ of the chain's bodies as a function of $\tau_{\PRB}(q_{\PRB}, \dot q_{\PRB})$.
In turn, the hybrid dynamics of a body chain takes the form 
\begin{equation}
        %\ddot q_\text{rfem} = \text{HD}(q_{\text{b}}, \dot q_{\text{b}}, \ddot q_{\text{b}}, %q_\text{rfem}, \dot q_\text{rfem}, \theta_{\text{P}}), \label{eq:hybrid-dynamics}
        \ddot q_{\PRB} = f_{\text{HybDyn}}(h, x, \theta_{\text{P}}), \label{eq:hybrid-dynamics}
\end{equation}
where $\theta_{\text{P}}$ denotes physical parameters \eg element length, mass, and inertia.
A prediction of the next state is obtained by rewriting \eqref{eq:hybrid-dynamics} in continuous state-space form $\dot h = f(h, x, \theta_{\text{P}})$ 
and resorting to numerically integration, reading
\begin{equation}
    h_{k+1} = \mathrm{ODESolve}(f, h_{k}, x_{k}, \Delta t), \label{eq:integration}
\end{equation}
where $\Delta t=t_{k+1} - t_k$ is the integration step size and $\mathrm{ODESolve}(\cdot)$ refers to a numerical integration method.

%Two main limitations of the analytical methods,  particularly the PRB, are expressiveness and slow integration time. The former stems from simple interaction forces between bodies, such as linear spring-damper \cite{wittbrodt2007RFEM}. The latter is caused by the stiffness of dynamics equations, which necessitate implicit integrators. Therefore, PRB-Net adopts discrete-time dynamics through NN. 

Two main weaknesses of PRB and other analytical models are their limited expressiveness and slow prediction times. Expressiveness of a PRB is limited by the choice of interaction force model. Commonly used interaction models such as linear spring-dampers require fine spatial discretization to yield tolerable approximations, which increases prediction times \cite{wittbrodt2007RFEM}. Computation is further slowed down as PRB dynamics are often stiff, which necessitates the use of implicit integration. PRB-Net improves prediction speed by modeling DLO dynamics in \emph{discrete-time} with neural networks. 

%=============================================================================================
\section{Method}
PRB-Net consists of three essential components as shown in the Fig. \ref{fig:FEIN}: a physics-informed encoder that maps initial observation $y_k$ and input $x_k$ into a hidden state $h_k$; a discrete-time dynamics network that propagates hidden states $h_k$ in time using inputs $x_k,\dots, x_{k+N}$; and an analytical decoder that maps hidden states and inputs into observations $y_k,\dots,y_{k+N}$. This decoder equals the forward kinematics of a DLO's PRB approximation.
% Recall that PRB approximates a DLO by a serial chain of rigid bodies connected via elastic joints. 

%By deploying the FK of the PRB approximation as a decoder, we enforce the dynamics network's latent state to represent the generalized coordinates of a PRB chain. 
%More precisely, the positions and velocities of the elastic joints.
%This approach not only grants physical interpretability to the model but also enables the learning of DLO dynamics from partial observations, facilitates the reconstruction of the DLO's shape, and paves the way for new avenues in physics-informed machine learning of DLO dynamics.

\begin{figure}
    \captionsetup{font=small}
    \centering
    \includegraphics[width=\columnwidth]{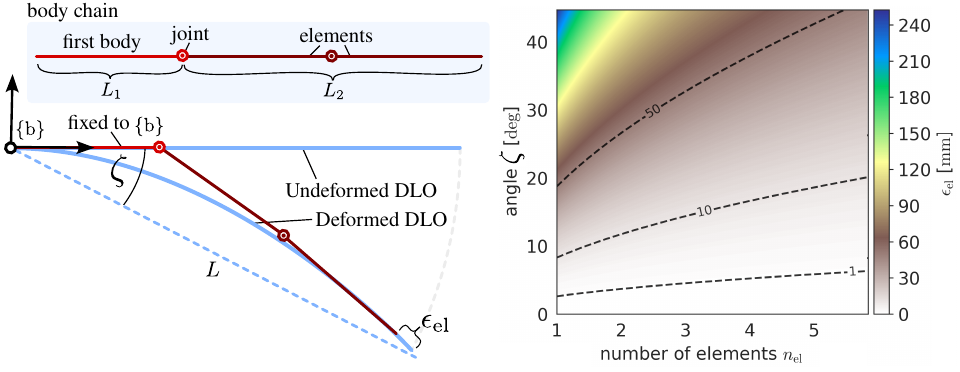}
    \caption{Errors arising in a rigid body chain approximation of a DLO with $\nseg=2$ and fixed element length. Left: If the chain's first body is fixed to $\{b\}$ then errors $\epsilon_{\mathrm{el}}$ arise in the prediction of the DLO's end position $p_{\text{e}}$. Right: Upper bound on $\epsilon_{\mathrm{el}}$ for $L=1920\,\mathrm{mm}$ caused by fixing the first body to $\{b\}$ and setting its length to $L_{\mathrm{el}}/2$ plotted over different element counts $\nseg$ and deflection angles $\zeta$.}
    \label{fig:dlo_approximation_errors}
    \vspace{-6mm}
\end{figure}

\subsection{Decoder}
The decoder in the \PRBN~is inspired by the kinematics description of a serial chain of rigid bodies. 
The decoder equals the FK as given by \eqref{eq:difpos} and \eqref{eq:difkin}. 
By using such a decoder, we ensure the interpretability of a neural network that models the DLO's dynamics, while maintaining a physically meaningful relationship between the hidden states. 

To derive an FK decoder, we specify the type of joint (\eg YZ Euler angles), provide the DLO's overall length, select the number of elements $\nseg$, and use \cite{wittbrodt2007RFEM} to obtain the initial element lengths. 

%\paragraph*{Discretization errors}
The uniform discretization method \cite{wittbrodt2007RFEM} assumes constant element lengths $L_{\mathrm{el}}$.
%while the length of the chain's first body is set to $L_{\mathrm{el}}/2$. %For small element counts, such a discretization spawns errors.
%
As illustrated in Fig.  \ref{fig:dlo_approximation_errors} (Left), fixing the first body to $\{\text{b}\}$ causes an error $\epsilon_{\mathrm{el}}$ in the \PRBN's prediction of the DLO end's position $p_{\text{e}}$. An upper bound for $\epsilon_{\mathrm{el}}$ is obtained using the law of cosines, writing
\begin{equation}
    \epsilon_{\mathrm{el}} < \sqrt{L^2 + L_1^2 - 2 L L_1\cos(\zeta)} - L_2,
    \label{eq:upper_bound}
\end{equation}
with the DLO's length $L=L_1+L_2$, the length of the first body $L_1$, the length of all other elements $L_2 $, and the deflection angle $\zeta$ as depicted in Fig.\ \ref{fig:dlo_approximation_errors}. 
%In turn, the discretization error $\epsilon_{\mathrm{el}}$ due to fixing the first body to $\{\mathrm{b}\}$ is upper bounded by \eqref{eq:upper_bound} if the deformed DLO's end point $p_{\text{e}}$ remains in a cone with angle $\zeta$ and slant height $L$. Fig.\,\ref{fig:dlo_approximation_errors} (Right) plots the upper bound on $\epsilon_{\mathrm{el}}$ for different $\zeta$ and $\nseg$. 

%\paragraph*{Learning the body lengths}
We reduce $\epsilon_{\mathrm{el}}$ by learning the body lengths $\theta_\text{el}\in \mathbb{R}^{n_{\text{el}}}_{>0}$ %
%= [L_1,\dots,L_{n_\text{el}+1}]$ 
along the \PRBN's network parameters. During the training of a \PRBN, we regularize the sum of estimated body lengths as detailed in Section \ref{sec:loss}.

\subsection{Encoder}
The encoder predicts the inverse kinematics of the serial chain. Internally, the encoder performs two key transformations when the data is fed to a multilayer perceptron (MLP) $q_\PRB = \Phi(p_{\text{e}}, q_{\text{b}})$. The transformations include: \emph{i)} compute the vector 
$p_{\mathrm{e},\mathrm{b}} = p_{\mathrm{e}} - p_{\mathrm{b}}$ to leverage that $q_\PRB$ is independent to translations of the DLO in space; \emph{ii)} transform the Euler angles $\Phi_b$ to $\sin(\cdot)$ and $\cos(\cdot)$ to avoid discontinuities in the angle parametrization.
The velocities $\dot q_\PRB$ are obtained via partial differentiation as
\begin{equation}
   \dot q_\PRB = \frac{\partial \Phi}{\partial p_\text{e}} \dot p_\text{e} + \frac{\partial \Phi}{\partial q_\text{b}} \dot q_\text{b}, \label{eq:dotq}
\end{equation}
which is straightforward to implement with automatic differentiation libraries such as JAX \cite{jax2018github}. Note, \eqref{eq:dotq} ensures that the velocity $\dot q_\PRB$ is the time derivative of $q_\PRB$, which is not the case if an MLP directly predicts $\dot q_\PRB$.
If $\text{dim}(q_\PRB)>\text{dim}([p_\text{e}, q_\text{b}])$, multiple chain configurations $q_\PRB$ yield the same pose of $\{\mathrm{e}\}$. In this case, learning inverse kinematics from observations is an ill-posed problem for deterministic networks \cite{JORDAN1992396}. 

\subsection{Dynamics Network}
The dynamics network unrolls the hidden state $h = [q_\PRB, \dot q_\PRB]$ -- being the generalized coordinates of the pseudo-rigid bodies --  over time. In turn, the dynamics network takes up a similar role as the analytical hybrid dynamics \eqref{eq:hybrid-dynamics} and \eqref{eq:integration}. Assuming that PRB is a suitable model for DLOs, these equations indicate that the only input needed for unrolling $h$ in time is $x=[q_{\text{b}}, \dot q_{\text{b}}, \ddot q_{\text{b}}]$. Therefore, the dynamics network $F$ is a neural network that receives as inputs $x_k$ and $h_k$ to predict $h_{k+1}$: $h_{k+1} = F(h_{k}, x_{k})$. The network $F$ can essentially have \emph{any} architecture; in this paper, we consider recurrent and residual NNs. 

While in our experiments, we attain precise estimates of $\ddot q_{\text{b}}$, future versions of \PRBN\ could draw inspiration from \cite{sanchez2020learning} and replace $\ddot q_{\text{b}}$ with past velocity observations. In addition, if one observes the robot's end-effector force $\tau_{\text{b}}$, then the dynamics network could learn \emph{forward dynamics} by replacing $x$ with the input $x^\prime=[q_{\text{b}}, \dot q_{\text{b}}, \tau_{\text{b}}]$.
\begin{figure}[t]
    \captionsetup{font=small} % footnotesize
    \centering
    \includegraphics[width=1.0\columnwidth, trim={0 -1cm 111.7cm 6cm}, clip]{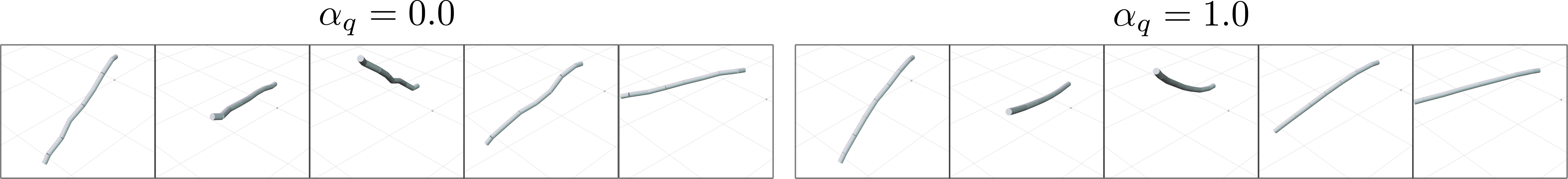}
    \includegraphics[width=1.0\columnwidth, trim={111.7cm 0 0 6cm}, clip]{images/dlo_shape_v5.pdf}
    \caption{Aluminium rod DLO shape predicted by a \PRBN~with $\nseg=7$. Top: $\alpha_q=0$. Bottom: $\alpha_q=1$.}
    \label{fig:shapes}
    \vspace{-6mm}
\end{figure}

\subsection{Loss Regularization} \label{sec:loss}
While we hypothesized that simultaneously training forward kinematics with dynamics may regularize $q_\PRB$, we observed that acquiring a physically plausible state representation in \PRBN~requires loss regularization. Therefore, we introduce a regularization term in the loss used to train \PRBN, writing
% \begin{equation}
% \resizebox{.9\hsize}{!}{\mathcal{L}_{\mathrm{FEI}} = \frac{1}{n_b N}\sum_{j=1}^{n_b} \sum_{k=1}^{N} w_k || y_{k,j} - \hat y_{k,j} ||_{W_y}^2 + \mathcal{L}_{\mathrm{reg}}(q_\PRB, \dot q_\PRB)},  \label{eq:loss} 
% \end{equation}
\begin{equation}
\mathcal{L} = \frac{1}{n_b N}\sum_{j=1}^{n_b} \sum_{k=1}^{N} w_k \| y_{k,j} - \hat y_{k,j} \|_{W_y}^2 + \mathcal{L}_{\mathrm{reg}}  \label{eq:loss} 
\end{equation}
with the regularization loss
\begin{equation}
     \mathcal{L}_{\mathrm{reg}} = 
     % \underbrace{\alpha_{q} q_{\text{rfem}, k,j}^{\transp} q_{\text{rfem},k,j}}_{\mathcal{L}_P} +
     \underbrace{\alpha_{q} \|q_{\PRB, k,j} \|_2^2}_{\mathcal{L}_P} +
     \underbrace{\alpha_{\dot q} \|\dot q_{\PRB, k,j}\|_2^2 }_{\mathcal{L}_K} + 
     \mathcal{L}_\text{FK}, \label{eq:regularization}
\end{equation}
the batch size $n_b$, the model's predictions $\hat y_{k,j}$, the weighting  matrix $W_y \in \mathbb{R}^{6\times 6}$, the state weight $w_k$, regularization parameters $\alpha_{q}$ and $\alpha_{\dot q}$, and additional kinematics regularization losses $\mathcal{L}_\text{FK}$. While the first two terms in \eqref{eq:regularization} can be simply seen as $\ell^2$ regularization of the hidden state $h$, we can look at these terms through the lens of the PRB method. 
%%%%%%
% Write something about the state weight
%%%%%%
In this regard, $\mathcal{L}_P$ in \eqref{eq:regularization} represents a potential energy of linear spring forces acting between the elements while $\mathcal{L}_K$ represents the chain's kinetic energy.
While the kinetic energy of the rigid body chain depends on the inertia matrix $M(q_\PRB)$, it is lower and upper bounded by positive real constants $\kappa_1 \leq M(q_\PRB) \leq \kappa_2$ \cite{siciliano2009modelling}. %\textbf{DLO shape}\hspace{0.3cm} 
%Accurate shape reconstruction from hidden states potentially enables safe manipulation of DLOs in environments with obstacles. 
We can reconstruct the DLO's shape from the \PRBN's hidden state $h$ using FK \eqref{eq:difpos}. As illustrated in Fig.  \ref{fig:shapes}, the chain's shape depends on the choice of $\alpha_{q}$. 
Finally, $\mathcal{L}_\text{FK}$ reads 
\begin{align}
    \mathcal{L}_\text{FK} = \alpha_L \left(\sum \theta_\text{el} - L\right) + \alpha_{\text{el}} \|\theta_\text{el} - \theta_\text{el}^{\prime}\|_2^2 + \alpha_{\text{eb}} \| \theta_{\text{eb}} - \theta_{\text{eb}}^{\prime}\|_2^2,
    \label{eq:L_el}
\end{align}
with $\theta_\text{el}^{\prime}$ denoting the initial element lengths obtained as in \cite{wittbrodt2007RFEM}, $\theta_{\text{eb}} $ denoting the position of the marker attached to the end of a DLO ${}^{\mathrm{b}}p_{\mathrm{e}, \mathrm{b}}$, and $\theta_{\text{eb}}^{\prime}$ denoting the same quantity obtained from prior calibration routines.
The first two terms in \eqref{eq:L_el} enable the \PRBN\ to reduce $\epsilon_{\mathrm{el}}$ by estimating the element lengths $\theta_\text{el}$ as shown in Fig.  \ref{fig:dlo_approximation_errors}.
To account for calibration errors in the calibration of ${}^{\mathrm{b}}p_{\mathrm{e}, \mathrm{b}}$, we introduce $\theta_\text{eb}$ to the kinematics dencoder model and also optimize its entries during training. %The third term in \eqref{eq:L_el} ensures that $\theta_\text{eb}$ remains close to $\theta_\text{eb}^{\prime}$.

\begin{figure}[t]
    \centering
    \captionsetup{font=small}
    \includegraphics[width=1.0\linewidth ,trim={0 2mm 0 2mm}, clip]{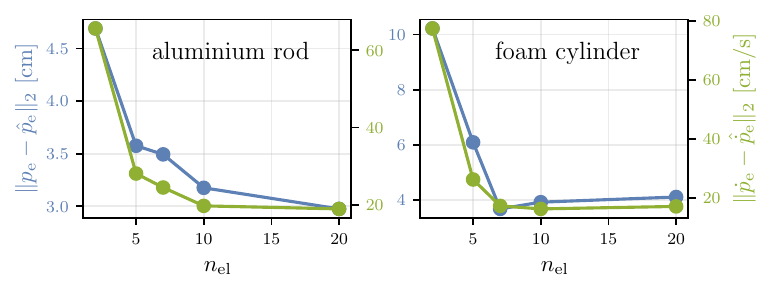}
    \caption{Influence of the element count $\nseg$ on the PRB-Net's prediction error of $\pe$ (blue) and $\dot p_{\text{e}}$ (red) for a $1\,\mathrm{s}$ prediction horizon.}
    \label{fig:element_count}
    \vspace{-6mm}
\end{figure}

%==============================================================================================
\section{Experimental evaluation}
\label{sec:result}
To evaluate the performance of different \PRBN, we learn to predict the motion of two physical DLOs with different properties that are rigidly connected to a Franka Panda robot arm. The two DLO's are: i) an aluminum rod as depicted in Fig.  \ref{fig:setup_and_method}, ii) a polyethylene foam cylinder as depicted in Fig. \ref{fig:noodle_comparison}. Kinematics parameters of DLOs (length, inner and outer diameters) are shown in Table \ref{tab:first_princ_params}. The project code and data are available at: \href{http://tinyurl.com/prb-networks}{tinyurl.com/prb-networks}. 

\subsection{Data}
We collected twelve different trajectories for the aluminum rod and eight trajectories for the foam cylinder. During the first half of each trajectory (about 15 seconds) the robot arm moved the DLO; during the second half, the robot arm rested while the DLO kept moving. For the aluminum rod, nine trajectories were used for training and validation, while for the foam cylinder, seven trajectories were collected. The trajectories were divided into rollouts of length $N=250$ using a rolling window approach, corresponding to one second of motion. Then, each rollout was split into 85\,\% for training and 15\,\% for validation.
% 
%Regarding model testing, we utilized three trajectories that had not been included in the training phase. 
The remaining trajectories of the DLO's were used for testing and divided into rollouts using a sliding window.

\subsection{Model Settings}
As the \PRBN's dynamics network, we used RNN with Gated Recurrent Units \cite{cho2014GRU} and ResNets \cite{he2016ResNet} -- and denote the resulting models as PRBN-RNN and PRBN-ResNet, respectively. 
As ``black-box'' baseline, we combine the same models with fully connected MLPs acting as encoder and decoder denoting the resulting models as RNN and ResNet. Note that the MLP encoder predicts the full state $h$ without enforcing $\dot q_\PRB = \mathrm{d}q_\PRB/\mathrm{d}t$. 
As a first-principles baseline, we employ the classical PRB method with the manually tuned parameters $\rho$, $E$ and $\eta$ provided in Table \ref{tab:first_princ_params}.
%following material properties for the aluminium rod: density 2710 $\text{kg/m}^{3}$, Young's modulus 51.5 GPa , normal damping coefficient 0.12 GNs/m.
The \PRBN\ models are trained using the loss function \eqref{eq:loss} while the baseline NNs use an $\ell^2$ loss. All models were implemented in JAX \cite{jax2018github} using the Equinox library \cite{kidger2021equinox}. Further details and hyperparameters are provided in the attached code repository.

\begin{figure}[t]
    \captionsetup{font=small}
    \centering
    \includegraphics[width=1.0\linewidth, trim={0 2mm 0 2mm}, clip]{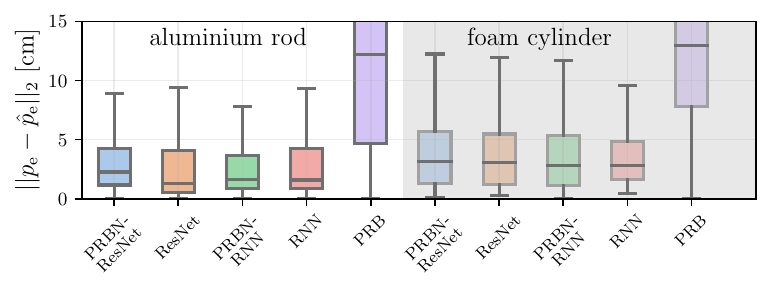}
    \caption{RMSE of PRBN- and black-box models for a rollout length of $N=250$ corresponding to a $1\,\mathrm{s}$ prediction horizon.}
    \label{fig:position-prediction-error}
    \vspace{-2mm}
\end{figure}

\begin{table}[]
    \captionsetup{font=small}
    \centering
    \setlength{\tabcolsep}{1.82pt}
    \begin{tabular}{ccccccc}
        \toprule
         DLO & $L$, m & $d_\mathrm{in}$, m & $d_\mathrm{out}$, m & $\rho,\ \text{kg/m}^{3}$ & $E$, Pa & $\eta,\ \text{Ns/m}$ \\
         \midrule
         Aluminum rod & 1.92 & $0.004$ & $0.006$ & 2710 & $5.15\cdot 10^{10}$ & $1.2 \cdot 10^8$ \\
         Foam cylinder & 1.90 & $ 0 $ & $0.06$ &105 & $1.8\cdot 10^6$ & $3.5 \cdot 10^5$\\
         \bottomrule
    \end{tabular}
    \caption{Physical parameters of a PRB model approximating the DLO dynamics with Young’s modulus $E$ and normal damping coefficient $\eta$. $d_\mathrm{in}$ and $d_\mathrm{out}$ are the inner and outer diameters, respectively. The PRB model with these parameters is used as a baseline for the experiments in Section \ref{sec:result}.}
    \label{tab:first_princ_params}
    \vspace{-6mm}
\end{table}

\subsection{Experimental Results}
\paragraph{Element Count}
Fig.  \ref{fig:element_count} depicts the RMSE of the PRBN-RNN model's predictions on the test data. Initially, increasing the number of elements substantially reduces the error. This observation aligns with the common knowledge that discretization-based methods, like PRB method, show improved model accuracy for increasing element counts. However, beyond seven elements, the marginal improvement in prediction accuracy does not justify the increased computational cost associated with model evaluation time, which is crucial for enabling efficient model-based control. Therefore, for the remainder of this paper, we adopt $n_\text{el}=7$.

\paragraph{Prediction Accuracy}
Fig.  \ref{fig:position-prediction-error} shows the prediction accuracy of the considered models  evaluated on test trajectories of the same length ($N=250$ corresponding to a prediction horizon of 1 second) as the training rollouts. Data-driven models significantly outperform the first-principles model (PRB). While physical interpretability often comes at the cost of performance (see PRB results), this is not the case for the PRB-Nets that provide essentially the same performance as black-box models, while being physically interpretable. The models more accurately predict the aluminum rod's motion than the foam cylinder, possibly due to variations in training data or material properties.

\begin{figure}[t]
    \captionsetup{font=small}
    \centering
    \includegraphics[width=1.0\linewidth ,trim={0 2mm 0 2mm}, clip]{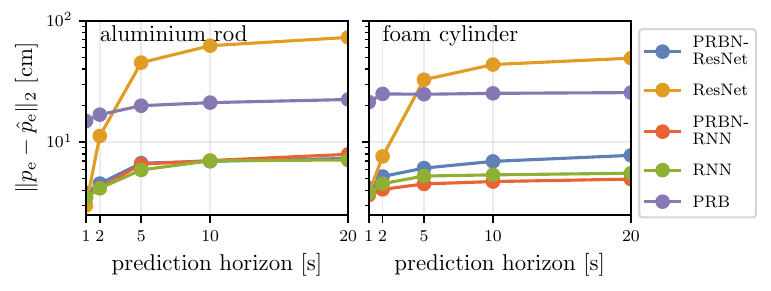}
    \caption{RMSE of $p_{\text{e}}$ for different prediction horizons.}
    \label{fig:pred_errors}
    \vspace{-3mm}
\end{figure}

\begin{figure}[t]
    \captionsetup{font=small}
    \centering
    \includegraphics[width=1.0\linewidth, trim={0.2cm 2.5mm 0 2.5mm}, clip]{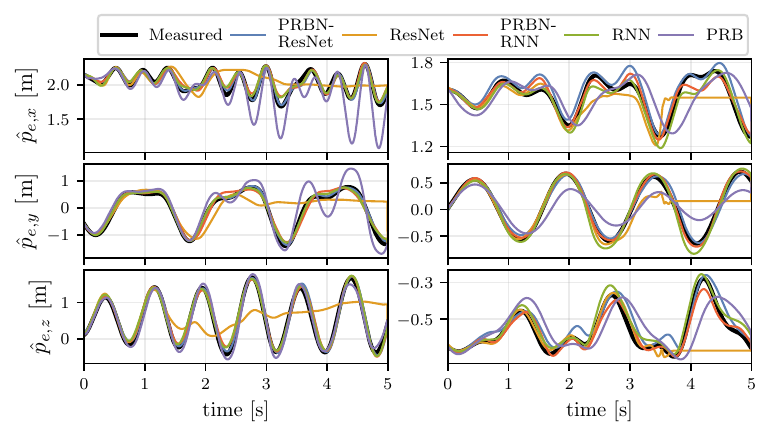}
    \caption{Models rollouts with a prediction horizon five times longer than during training. Left: Aluminum rod. Right: Foam cylinder.}
    \label{fig:long_output_pred}
    \vspace{-5mm}
\end{figure}

All the models yield stable predictions during rollouts that extend significantly beyond $N$, as depicted in Fig. \ref{fig:pred_errors} and \ref{fig:long_output_pred}. However, the ResNet exhibits a noticeable drop in prediction accuracy as the prediction length exceeds $N$. In contrast, the PRBN-ResNet does not show this issue.
Figure \ref{fig:pred_errors} shows the model's RMSE of the prediction of $p_{\text{e}}$ for both DLOs. Our findings indicate that the \PRBN\ and  models incorporating black-box encoders and decoders show similar performance. 
Among models with a black-box decoder, the RNN has a smaller RMSE than the ResNet for all rollouts except for the one of length $N$. Additionally, the \PRBN\ exhibits slightly higher RMSEs compared to the RNN at $20N$.
These observations align with literature on position encodings \cite{dufter2022positioninfo} that suggest that resorting to a low dimensional latent space whose dimensionality is determined by physics reduces the model's expressiveness in favor of physical interpretability.

\paragraph{Training and Model Evaluation Times}

In terms of training time, \PRBN\ is slower (about 1.5 times) to train than standard neural networks as the nonlinear forward kinematics and the physics-informed encoder require gradient computations during the forward pass as shown in Fig. \ref{fig:FEIN}.
Since the trained models may be used for model-based control, we analyzed the prediction times for one second-ahead prediction of the best data-driven models and of analytical PRB models. The PRBs analytical continuous-time dynamics were integrated using a constant step-size implicit 3rd order Runge-Kutta integrator. 

Table \ref{tab:inference_time} shows that the PRBN-RNN and RNN models have similar evaluation times, with the latter being slightly faster. However, even with just a two-element discretization, the evaluation time for the PRB method is at least two orders of magnitude slower. While integration times could be reduced by selecting a dedicated integrator, they will likely not match the discrete-time models like PRB-Net which requires only a single function evaluation. %This result highlights the ...

\begin{figure}[t]
    \centering
    \captionsetup{font=small}
    \includegraphics[width=0.45\textwidth]{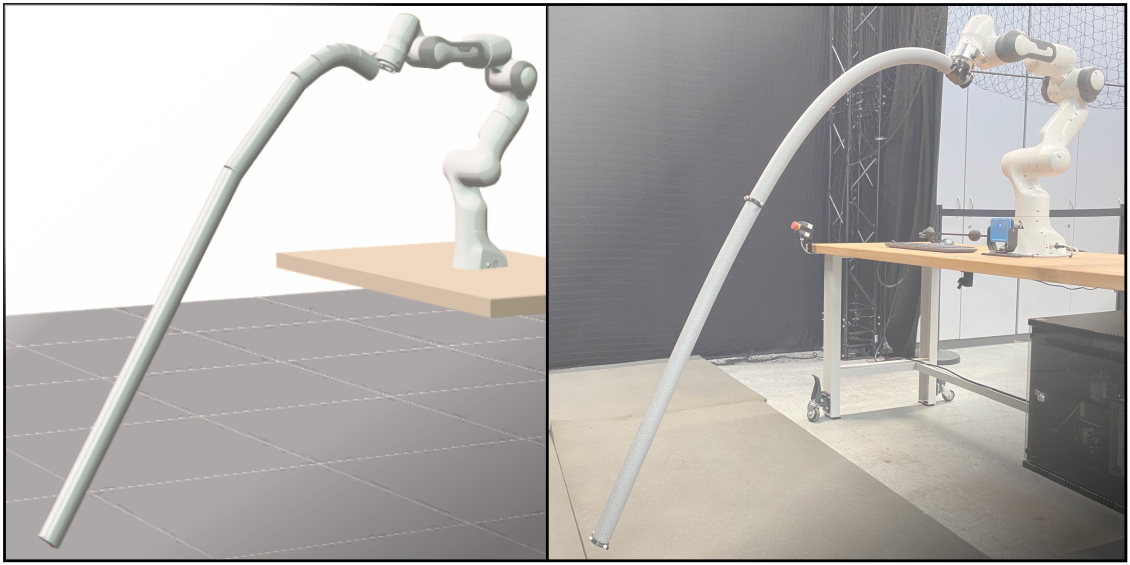} % 0.48
    \caption{Left: Shape of a foam cylinder predicted by a \PRBN~with $\nseg=7$ and the element length being learned alongside other parameters. Right: Foam cylinder actuated by a robot arm.}
    \label{fig:noodle_comparison}
    %\vspace{-1mm}
\end{figure}

\paragraph{Physical interpretability}
\PRBN s possess a physically meaningful state representation compared to black-box models such as RNNs and ResNets. In this regard, \PRBN s hidden state represents elastic joint positions and velocities between rigid bodies. Therefore, one can approximately reconstruct the DLO's shape via forward kinematics \eqref{eq:difpos} from a partial observation $y$ and an input $x$, as shown in Fig. \ref{fig:noodle_comparison}. %This property is especially valuable in manipulation tasks involving obstacles.

\begin{table}[t]
    \captionsetup{font=small}
    \centering
    \setlength{\tabcolsep}{3.3pt}
    \begin{tabular}{ccccccc}
        \toprule
        & RNN & PRBN-RNN &  PRB-2 &PRB-5 & PRB-7 & PRB-10\\
        \midrule
        \rowcolor[HTML]{ECF4FF}
        \begin{tabular}[c]{@{}l@{}} Pred.\ \\ time [s]\end{tabular} & $4.3\cdot 10^{-3}$ & $5.7\cdot 10^{-3}$ & $1.5$ &  $2.3$ & $3.1$ &  $4.3$ \\
        \bottomrule
    \end{tabular}
    \caption{Model prediction times for a one second-ahead prediction. For the analytical PRB method several spatial discretizations were considered \eg $\nseg=2$ for ``PRB-2''.}
    \label{tab:inference_time}
    \vspace{-3mm}
\end{table}

%==============================================================================================
\section{Conclusion}
\label{sec:conclusion}
We presented the first data-driven DLO dynamics model capable of accurately learning the motion of a DLO from partial observations while also acquiring a physically interpretable hidden state representation. 
This model was obtained by first deriving the forward kinematics of a rigid body chain approximating the DLO. At the first step of a rollout, the DLO's hidden state is reconstructed from partial observations of the DLO's start and end frames using an MLP encoder. In the subsequent time steps, the hidden state is unrolled through time using a neural network that effectively learns the system's dynamics. All networks are trained jointly with the kinematics parameters while we regularize the hidden state to ensure that it acquires a physically plausible hidden state representation.

Our model successfully predicted the position and velocity of a 1.92\,m long aluminum rod and a 1.90\,m long polyethylene foam cylinder. Despite the endpoints of the aluminium rod moving up to 1.5\,m from their equilibrium position, the trained \PRBN\ achieved a sub-decimeter RMSE in a twenty-second simulation.

\addtolength{\textheight}{-2.1cm}   % This command serves to balance the column lengths

%%%%%%%%%%%%%%%%%%%%%%%%%%%%%%%%%%%%%%%%%%%%%%%%%%%%%%%%%%%%%%%%%%%%%%%%%%%%%%%%

%%%%%%%%%%%%%%%%%%%%%%%%%%%%%%%%%%%%%%%%%%%%%%%%%%%%%%%%%%%%%%%%%%%%%%%%%%%%%%%%



\section*{Acknowledgment}

The authors thank Sebastian Giedyk for helping with the setup, Henrik Hose for his inputs on ROS, and Wilm Decré for his valuable feedback on the manuscript.

%%%%%%%%%%%%%%%%%%%%%%%%%%%%%%%%%%%%%%%%%%%%%%%%%%%%%%%%%%%%%%%%%%%%%%%%%%%%%%%%
\newpage
\bibliographystyle{IEEEtransBST/IEEEtran}
\bibliography{bibliography.bib}

\begin{thebibliography}{10}
\providecommand{\url}[1]{#1}
\csname url@rmstyle\endcsname
\providecommand{\newblock}{\relax}
\providecommand{\bibinfo}[2]{#2}
\providecommand\BIBentrySTDinterwordspacing{\spaceskip=0pt\relax}
\providecommand\BIBentryALTinterwordstretchfactor{4}
\providecommand\BIBentryALTinterwordspacing{\spaceskip=\fontdimen2\font plus
\BIBentryALTinterwordstretchfactor\fontdimen3\font minus \fontdimen4\font\relax}
\providecommand\BIBforeignlanguage[2]{{%
\expandafter\ifx\csname l@#1\endcsname\relax
\typeout{** WARNING: IEEEtran.bst: No hyphenation pattern has been}%
\typeout{** loaded for the language `#1'. Using the pattern for}%
\typeout{** the default language instead.}%
\else
\language=\csname l@#1\endcsname
\fi
#2}}

\bibitem{sanchez2018robotic}
J.~Sanchez, J.-A. Corrales, B.-C. Bouzgarrou, and Y.~Mezouar, ``Robotic manipulation and sensing of deformable objects in domestic and industrial applications: a survey,'' \emph{The International Journal of Robotics Research}, vol.~37, no.~7, pp. 688--716, 2018.

\bibitem{jimenez2012survey}
P.~Jim{\'e}nez, ``Survey on model-based manipulation planning of deformable objects,'' \emph{Robotics and computer-integrated manufacturing}, vol.~28, no.~2, pp. 154--163, 2012.

\bibitem{wang2015online}
W.~Wang, D.~Berenson, and D.~Balkcom, ``An online method for tight-tolerance insertion tasks for string and rope,'' in \emph{2015 IEEE International Conference on Robotics and Automation (ICRA)}.\hskip 1em plus 0.5em minus 0.4em\relax IEEE, 2015, pp. 2488--2495.

\bibitem{yoshida2015simulation}
E.~Yoshida, K.~Ayusawa, I.~G. Ramirez-Alpizar, K.~Harada, C.~Duriez, and A.~Kheddar, ``Simulation-based optimal motion planning for deformable object,'' in \emph{2015 IEEE international workshop on advanced robotics and its social impacts (ARSO)}.\hskip 1em plus 0.5em minus 0.4em\relax IEEE, 2015, pp. 1--6.

\bibitem{yan2020selfsupervised}
M.~Yan, Y.~Zhu, N.~Jin, and J.~Bohg, ``Self-supervised learning of state estimation for manipulating deformable linear objects,'' \emph{IEEE Robotics and Automation Letters}, vol.~5, no.~2, pp. 2372--2379, 2020.

\bibitem{pai2002strands}
D.~K. Pai, ``Strands: Interactive simulation of thin solids using cosserat models,'' in \emph{Computer graphics forum}, vol.~21.\hskip 1em plus 0.5em minus 0.4em\relax Wiley Online Library, 2002, pp. 347--352.

\bibitem{saha2007manipulation}
M.~Saha and P.~Isto, ``Manipulation planning for deformable linear objects,'' \emph{IEEE Transactions on Robotics}, vol.~23, no.~6, 2007.

\bibitem{verl2019}
A.~Verl, A.~Valente, S.~Melkote, C.~Brecher, E.~Ozturk, and L.~T. Tunc, ``Robots in machining,'' \emph{CIRP Annals}, vol.~68, no.~2, pp. 799--822, 2019.

\bibitem{lutter2019deep}
M.~Lutter, C.~Ritter, and J.~Peters, ``Deep lagrangian networks: Using physics as model prior for deep learning,'' in \emph{International Conference on Learning Representations}, 2019.

\bibitem{cranmer2019lagrangian}
M.~Cranmer, S.~Greydanus, S.~Hoyer, P.~Battaglia, D.~Spergel, and S.~Ho, ``Lagrangian neural networks,'' in \emph{ICLR 2020 Workshop on Integration of Deep Neural Models and Differential Equations}, 2019.

\bibitem{rath2022using}
L.~Rath, A.~R. Geist, and S.~Trimpe, ``Using physics knowledge for learning rigid-body forward dynamics with gaussian process force priors,'' in \emph{Conference on Robot Learning}.\hskip 1em plus 0.5em minus 0.4em\relax PMLR, 2022, pp. 101--111.

\bibitem{nguyen2010using}
D.~Nguyen-Tuong and J.~Peters, ``Using model knowledge for learning inverse dynamics,'' in \emph{2010 IEEE International Conference on Robotics and Automation}, 2010, pp. 2677--2682.

\bibitem{lutter2023combining}
M.~Lutter and J.~Peters, ``Combining physics and deep learning to learn continuous-time dynamics models,'' \emph{The International Journal of Robotics Research}, vol.~42, no.~3, pp. 83--107, 2023.

\bibitem{spillmann2007corde}
J.~Spillmann and M.~Teschner, ``Corde: Cosserat rod elements for the dynamic simulation of one-dimensional elastic objects,'' in \emph{Proceedings of the 2007 ACM SIGGRAPH/Eurographics symposium on Computer animation}, 2007, pp. 63--72.

\bibitem{drucker2023application}
S.~Dr{\"u}cker and R.~Seifried, ``Application of stable inversion to flexible manipulators modeled by the absolute nodal coordinate formulation,'' \emph{GAMM-Mitteilungen}, vol.~46, no.~1, 2023.

\bibitem{kim2011fast}
J.~Kim and N.~S. Pollard, ``Fast simulation of skeleton-driven deformable body characters,'' \emph{ACM Transactions on Graphics (TOG)}, vol.~30, no.~5, pp. 1--19, 2011.

\bibitem{bergou2008discrete}
M.~Bergou, M.~Wardetzky, S.~Robinson, B.~Audoly, and E.~Grinspun, ``Discrete elastic rods,'' in \emph{ACM SIGGRAPH 2008}, 2008, pp. 1--12.

\bibitem{wittbrodt2007RFEM}
E.~Wittbrodt, I.~Adamiec-W{\'o}jcik, and S.~Wojciech, \emph{Dynamics of flexible multibody systems: rigid finite element method}.\hskip 1em plus 0.5em minus 0.4em\relax Springer Science \& Business Media, 2007.

\bibitem{boyer2024CosseratRodInteg}
F.~Boyer, A.~Gotelli, P.~Tempel, V.~Lebastard, F.~Renda, and S.~Briot, ``Implicit time-integration simulation of robots with rigid bodies and cosserat rods based on a newton–euler recursive algorithm,'' \emph{IEEE Transactions on Robotics}, vol.~40, pp. 677--696, 2024.

\bibitem{lang2011multi}
H.~Lang, J.~Linn, and M.~Arnold, ``Multi-body dynamics simulation of geometrically exact cosserat rods,'' \emph{Multibody System Dynamics}, vol.~25, no.~3, pp. 285--312, 2011.

\bibitem{webster2010design}
R.~J. Webster~III and B.~A. Jones, ``Design and kinematic modeling of constant curvature continuum robots: A review,'' \emph{The International Journal of Robotics Research}, vol.~29, no.~13, pp. 1661--1683, 2010.

\bibitem{stella2022underwatertentacle}
F.~Stella, N.~Obayashi, C.~D. Santina, and J.~Hughes, ``An experimental validation of the polynomial curvature model: Identification and optimal control of a soft underwater tentacle,'' \emph{IEEE Robotics and Automation Letters}, vol.~7, no.~4, pp. 11\,410--11\,417, 2022.

\bibitem{stella2022piecewise}
F.~Stella, Q.~Guan, C.~Della~Santina, and J.~Hughes, ``Piecewise affine curvature model: a reduced-order model for soft robot-environment interaction beyond pcc,'' in \emph{2023 IEEE International Conference on Soft Robotics (RoboSoft)}, 2023, pp. 1--7.

\bibitem{tanaka2022continuum-body-pose}
K.~Tanaka, Y.~Minami, Y.~Tokudome, K.~Inoue, Y.~Kuniyoshi, and K.~Nakajima, ``Continuum-body-pose estimation from partial sensor information using recurrent neural networks,'' \emph{IEEE Robotics and Automation Letters}, vol.~7, no.~4, pp. 11\,244--11\,251, 2022.

\bibitem{tariverdi2021recurrent}
A.~Tariverdi, V.~K. Venkiteswaran, M.~Richter, O.~J. Elle, J.~T{\o}rresen, K.~Mathiassen, S.~Misra, and {\O}.~G. Martinsen, ``A recurrent neural-network-based real-time dynamic model for soft continuum manipulators,'' \emph{Frontiers in Robotics and AI}, vol.~8, p. 631303, 2021.

\bibitem{yang2021learning}
Y.~Yang, J.~A. Stork, and T.~Stoyanov, ``Learning to propagate interaction effects for modeling deformable linear objects dynamics,'' in \emph{2021 IEEE International Conference on Robotics and Automation (ICRA)}.\hskip 1em plus 0.5em minus 0.4em\relax IEEE, 2021, pp. 1950--1957.

\bibitem{preiss2022trackingfast}
J.~A. Preiss, D.~Millard, T.~Yao, and G.~S. Sukhatme, ``Tracking fast trajectories with a deformable object using a learned model,'' in \emph{2022 International Conference on Robotics and Automation (ICRA)}, 2022, pp. 1351--1357.

\bibitem{hochreiter1997lstm}
S.~Hochreiter and J.~Schmidhuber, ``Long short-term memory,'' \emph{Neural computation}, vol.~9, no.~8, pp. 1735--1780, 1997.

\bibitem{scarselli2008graph}
F.~Scarselli, M.~Gori, A.~C. Tsoi, M.~Hagenbuchner, and G.~Monfardini, ``The graph neural network model,'' \emph{IEEE transactions on neural networks}, vol.~20, no.~1, pp. 61--80, 2008.

\bibitem{lynch2017modernrobotics}
K.~M. Lynch and F.~C. Park, \emph{Modern robotics}.\hskip 1em plus 0.5em minus 0.4em\relax Cambridge University Press, 2017.

\bibitem{huynh2009metrics}
D.~Q. Huynh, ``Metrics for 3d rotations: Comparison and analysis,'' \emph{Journal of Mathematical Imaging and Vision}, vol.~35, pp. 155--164, 2009.

\bibitem{bregier2021deep}
R.~Br{\'e}gier, ``Deep regression on manifolds: a 3d rotation case study,'' in \emph{2021 International Conference on 3D Vision (3DV)}.\hskip 1em plus 0.5em minus 0.4em\relax IEEE, 2021, pp. 166--174.

\bibitem{kim2012lie}
J.~Kim, ``Lie group formulation of articulated rigid body dynamics,'' Technical report, Carnegie Mellon University, Tech. Rep., 2012.

\bibitem{jax2018github}
\BIBentryALTinterwordspacing
J.~Bradbury, R.~Frostig, P.~Hawkins, M.~J. Johnson, C.~Leary, D.~Maclaurin, G.~Necula, A.~Paszke, J.~Vander{P}las, S.~Wanderman-{M}ilne, and Q.~Zhang, ``{JAX}: composable transformations of {P}ython+{N}um{P}y programs,'' 2018. [Online]. Available: \url{http://github.com/google/jax}
\BIBentrySTDinterwordspacing

\bibitem{JORDAN1992396}
M.~I. Jordan, ``Constrained supervised learning,'' \emph{Journal of Mathematical Psychology}, vol.~36, no.~3, pp. 396--425, 1992.

\bibitem{sanchez2020learning}
A.~Sanchez-Gonzalez, J.~Godwin, T.~Pfaff, R.~Ying, J.~Leskovec, and P.~Battaglia, ``Learning to simulate complex physics with graph networks,'' in \emph{International conference on machine learning}.\hskip 1em plus 0.5em minus 0.4em\relax PMLR, 2020, pp. 8459--8468.

\bibitem{siciliano2009modelling}
B.~Siciliano, L.~Sciavicco, L.~Villani, and G.~Oriolo, ``Modelling, planning and control,'' \emph{Advanced Textbooks in Control and Signal Processing. Springer,}, 2009.

\bibitem{cho2014GRU}
K.~Cho, B.~Van~Merri{\"e}nboer, C.~Gulcehre, D.~Bahdanau, F.~Bougares, H.~Schwenk, and Y.~Bengio, ``Learning phrase representations using rnn encoder-decoder for statistical machine translation,'' \emph{arXiv preprint arXiv:1406.1078}, 2014.

\bibitem{he2016ResNet}
K.~He, X.~Zhang, S.~Ren, and J.~Sun, ``Deep residual learning for image recognition,'' in \emph{Proceedings of the IEEE conference on computer vision and pattern recognition}, 2016, pp. 770--778.

\bibitem{kidger2021equinox}
P.~Kidger and C.~Garcia, ``{E}quinox: neural networks in {JAX} via callable {P}y{T}rees and filtered transformations,'' \emph{Differentiable Programming workshop at Neural Information Processing Systems 2021}, 2021.

\bibitem{dufter2022positioninfo}
\BIBentryALTinterwordspacing
P.~Dufter, M.~Schmitt, and H.~Schütze, ``{Position Information in Transformers: An Overview},'' \emph{Computational Linguistics}, vol.~48, no.~3, pp. 733--763, 09 2022. [Online]. Available: \url{https://doi.org/10.1162/coli\_a\_00445}
\BIBentrySTDinterwordspacing

\end{thebibliography}

\end{document}